\documentclass[sigconf]{acmart}

\usepackage{caption}
\usepackage{amsmath,amssymb,amsfonts}
\usepackage{algorithmic}
\usepackage{graphicx}
\usepackage{textcomp}
\usepackage{xcolor}
\usepackage{graphicx}
\usepackage{diagbox} 
\usepackage{verbatim}
\usepackage{comment}
\usepackage[utf8]{inputenc}
\usepackage[english]{babel}
\usepackage{float}
\usepackage[font=footnotesize]{subfig}

\AtBeginDocument{%
  \providecommand\BibTeX{{%
    \normalfont B\kern-0.5em{\scshape i\kern-0.25em b}\kern-0.8em\TeX}}}

\begin{document}

\title{F-Cooper: Feature based Cooperative Perception for Autonomous Vehicle Edge Computing System Using 3D Point Clouds}



\author{Qi Chen, Xu Ma, Sihai Tang, Jingda Guo, Qing Yang, Song Fu}
\affiliation{%
  \textit{\institution{Department of Computer Science and Engineering}} 
  \textit{\institution{University of North Texas, Denton, TX, USA}}
}
\email{{QiChen, XuMa, SihaiTang, JingdaGu}@my.unt.edu, {Qing.Yang, Song.Fu}@unt.edu}

\begin{abstract}

Autonomous vehicles are heavily reliant upon their sensors to perfect the perception of surrounding environments, however, with the current state of technology, the data which a vehicle uses is confined to that from its own sensors. 
%
Data sharing between vehicles and/or edge servers is limited by the available network bandwidth and the stringent real-time constraints of autonomous driving applications. %
To address these issues, we propose a point cloud feature based cooperative perception framework (F-Cooper) for connected autonomous vehicles to achieve a better object detection precision. 
Not only will feature based data be sufficient for the training process, we also use the features' intrinsically small size to achieve real-time edge computing, without running the risk of congesting the network. 
Our experiment results show that by fusing features, we are able to achieve a better object detection result, around 10\% improvement for detection within 20 meters and 30\% for further distances, as well as achieve faster edge computing with a low communication delay, requiring 71 milliseconds in certain feature selections.
To the best of our knowledge, we are the first to introduce feature-level data fusion to connected autonomous vehicles for the purpose of enhancing object detection and making real-time edge computing on inter-vehicle data feasible for autonomous vehicles.

\end{abstract}


\maketitle


\section{Introduction}
Connected autonomous vehicles (CAV) provide a promising solution to improving road safety. This relies on vehicles being able to perceive road conditions and detect objects precisely in real-time. However, accurate and real-time perception is challenging in the field. It involves processing high-volume and continuous data streams from various sensors with strict timing requirements. Moreover, the perception accuracy of a vehicle is often affected by the limited view and scope of the sensors. 
Edge computing can help CAVs achieve better situational awareness via combining and processing information collected from multiple CAVs with more powerful machine learning technologies~\cite{wang2018cavbench,shi2016edge}.
The ultimate goal of integrating edge computing and CAVs is to efficiently analyze massive amount of data in real time under limited network bandwidth.  

An autonomous vehicle edge computing system consists of three layers: vehicle, edge, and cloud~\cite{vedge}.
Each autonomous vehicle is equipped with  onboard edge device(s) that integrates the needed functional modules for autonomous driving, including localization, perception, path planning, and vehicle control.
Autonomous vehicles can communicate with roadside edge servers, and eventually reach the cloud through wireless networks, e.g., the dedicated short range communication (DSRC)~\cite{dsrc}, 5G or millimeter-wave communication technologies~\cite{va2016millimeter}. 
This provides a perfect opportunity to develop a cooperative perception system in which vehicles exchange their data with nearby edge servers. It is here that data are fused and processed to further extend the individual vehicle's perception range; beyond line-of-sight and beyond field-of-view.  

\subsection{Motivation}

\begin{figure}[!t]
    \centering
    \includegraphics[width=0.95\linewidth]{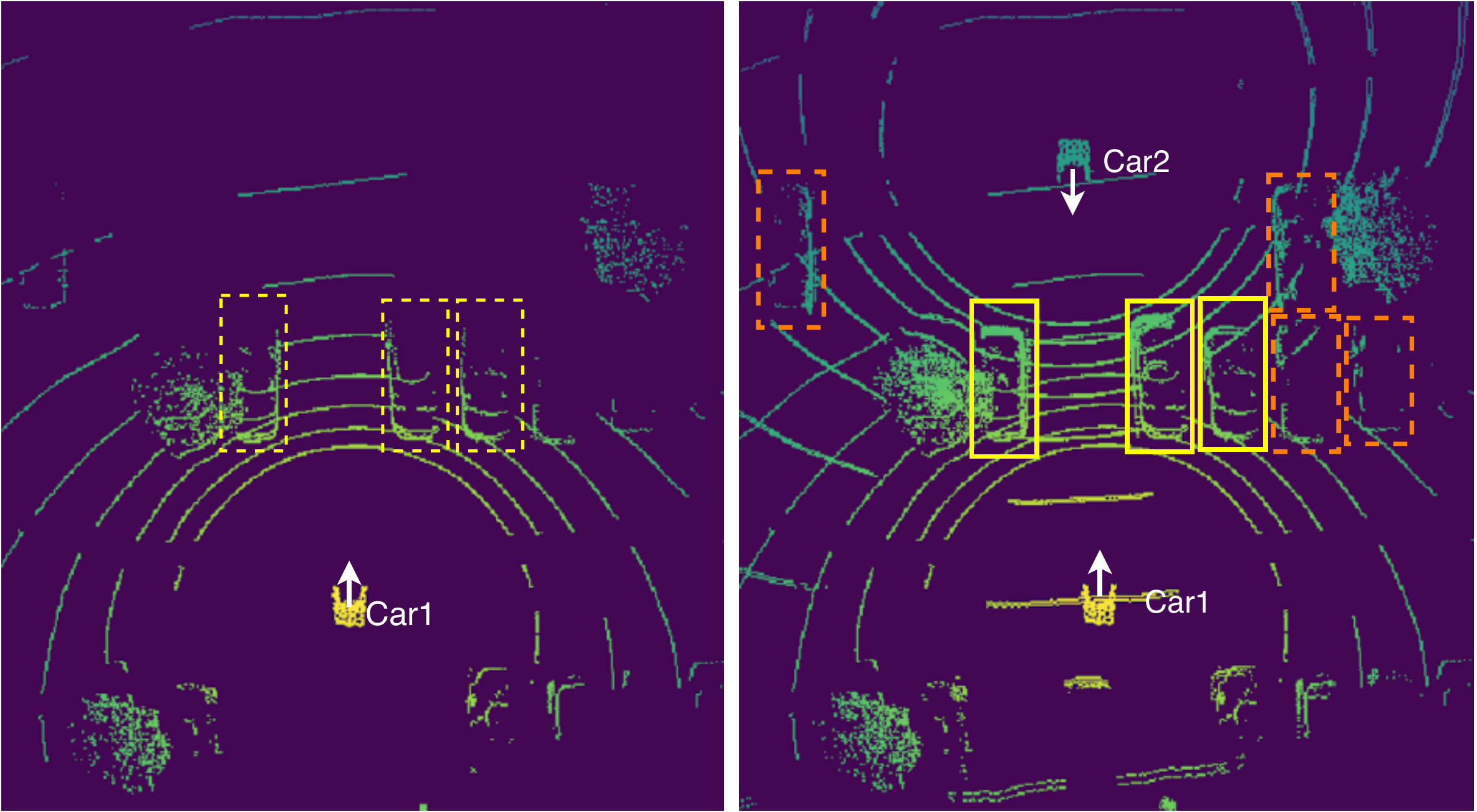}
    \caption{Occlusion and truncation situations naturally occur in point clouds data. In the left LiDAR image,  only three vehicles (yellow boxes) are recognized by Car 1. When it cooperatively detects with Car 2, four more vehicles (either occluded or truncated) are detected, as shown in red boxes in the right image, which are not detected using its own data.}
    \vspace{-20pt}
    \label{fig:acclusion}
\end{figure}

Having a single source of data input for autonomous vehicles is risky in real-world environments, as sensors are just another component of the vehicle that is susceptible to failure. 
In addition, sensors are also limited by their physical capabilities such as scan frequency, range, and resolution. 
Perception gets even worse when sensors are occluded, as shown in Fig.~\ref{fig:acclusion}.

Among related works on cooperative perception for autonomous vehicles~\cite{rauch2012car2x,cho2014multi}, we find that their main focus is on improving the individual vehicle's precision, overlooking benefits from cooperative perception. Potential issues involved in cooperative perception, such as accuracy of local perception results, impact on networks, format of data to be exchanged, and data fusion on edge servers, are not addressed. 
When it comes to 3D object detection, Lidar is one of the most important components of autonomous driving vehicles as it generates 3D point clouds to capture the 3D structures of scenes. This data gives precise location in 3D space with respect to the LiDAR, and by extension, the car. 

Based on our best acknowledge, the state of the art 3D object detection precision based on monocular LiDAR (Light Detection and Ranging) data comes from VoxelNet \cite{zhou2018voxelnet}, SECOND \cite{yan2018second} and PointRCNN~\cite{shi2018pointrcnn}, etc. For example, PointRCNN achieves 75.76\% mAP (mean average precision) on the KITTI moderate benchmark~\cite{geiger2012we}, and 85.94\%, 68.32\% on easy and hard benchmarks, respectively.
That implies the simple fusion of object detection results from different cars would yield errors.
Although fusing raw LiDAR data from two vehicles can improve the car detection precision ~\cite{qi2019cooper}, it is challenging to send the huge amount of LiDAR data generated by autonomous vehicles in real time.
Solutions that increase vehicle's perception precision as well as maintaining or reducing computational complexity and turnaround time are rare in the literature.

\subsection{Proposed Solution}
We propose a method that improves the autonomous vehicle's detection precision without introducing much computational overhead.
An useful insight is that modern object detection techniques for autonomous vehicles, both image based \cite{ren2015faster,liu2016ssd} and 3D LiDAR data based \cite{yan2018second,qi2018frustum}, commonly adopt a convolutional neural network (CNN) \cite{simonyan2014very,he2016deep} to process raw data, and leverage a region proposal network (RPN) \cite{ren2015faster} to detect objects. 
We argue that the capacity of feature maps is not fully explored, especially for 3D LiDAR data generated on autonomous vehicles, as the feature maps are used for object detection only by single vehicles.

To this end, we introduce a {\em feature based cooperative perception (F-Cooper) framework} that realizes an end-to-end 3D object detection leveraging feature-level fusion to improve
detection precision. 
%
%
Our F-Cooper framework supports two different fusion schemes: voxel feature fusion and spatial feature fusion.
While the former achieves almost the same detection precision improvement when compared to the raw-data level fusion solution~\cite{qi2019cooper}, the latter offers the ability to dynamically adjust the size of feature maps to be transmitted.
A unique characteristic of F-Cooper is that it can be deployed and executed on in-vehicle and roadside edge systems.

Aside from being able to improve detection precision, data needed for feature fusion is only one hundredth of the size of the original data.
For a typical LiDAR sensor, each LiDAR frame contains about 100,000 points, which is about 4 MB. Such huge amount data would become a severe burden for any existing wireless network infrastructure. In stark contrast to the large volume of raw LiDAR data, the size of features generated by a CNN can be as low as 200 Kb after compression techniques is applied. 
Empirical evidences from our experiments demonstrate that transmitting these features only takes dozens of milliseconds, which makes real-time edge computing feasible. 
Such a negligible cost also enables feature-level fusion to become an ideal choice for connected autonomous vehicles to improve detection precision while keeping a reasonable communication time.


%


\subsection{Main Contributions}
To the best of our knowledge, we are the first to propose feature map fusion based 3D object detection for connected autonomous vehicles on the edge. Through our experimentation and analysis, we have proved that not only does feature fusion provide an enhanced perception, it also allows for data to be compressed without losing detection value. 

With this data compression factor, we are able to state with confidence that our feature fusion strategies are suited for autonomous vehicles On-Edge. Due to the fact that vehicles have a limited amount of computational resources on-board, we look towards the edge for more powerful and reliable computational power. Should an autonomous vehicle require extra perception, it only needs to send its compressed feature data and receive either a detection result or a compressed, fused feature map, or even both. By cutting out the computational step, we are effectively pushing the heavy workload onto the edge and mitigating any downsides to data sharing. 

As proven in our experiments, both the data size and network transmission time are small enough that even in the most congested areas, vehicle data transmission will still be smooth. Both voxel feature fusion and spatial feature fusion perform better than the baseline for single vehicles without fusion, both the fusion and non-fusion baseline are derived from the same detection model. While spatial feature fusion data can be dynamically adjusted for a smaller compression size than voxel feature fusion data, the latter is capable of detection improvement on par with raw-data level fusion \cite{qi2019cooper}. With each strategy holding its own special advantages, we believe that our F-Cooper framework makes a substantial contribution that allows improvement no matter whether deployed in-vehicle or on roadside edge systems.

The remainder of this paper is organized as follows. Section 2 analyzes the properties of features to see if features are fit for fusion. Section 3 explains how our feature based methods work and outlines their place in F-Cooper. Section 4 tests our methods in fusion scenarios and evaluates suitability for on-edge deployment. Section 5 and 6 discuss previous works and related studies. Finally, Section 7 concludes this paper.


\section{Towards Feature based Fusion of Vehicle Data}


\subsection{Convolutional Feature Maps}

\begin{figure}
    \centering
    \includegraphics[width=0.9\linewidth]{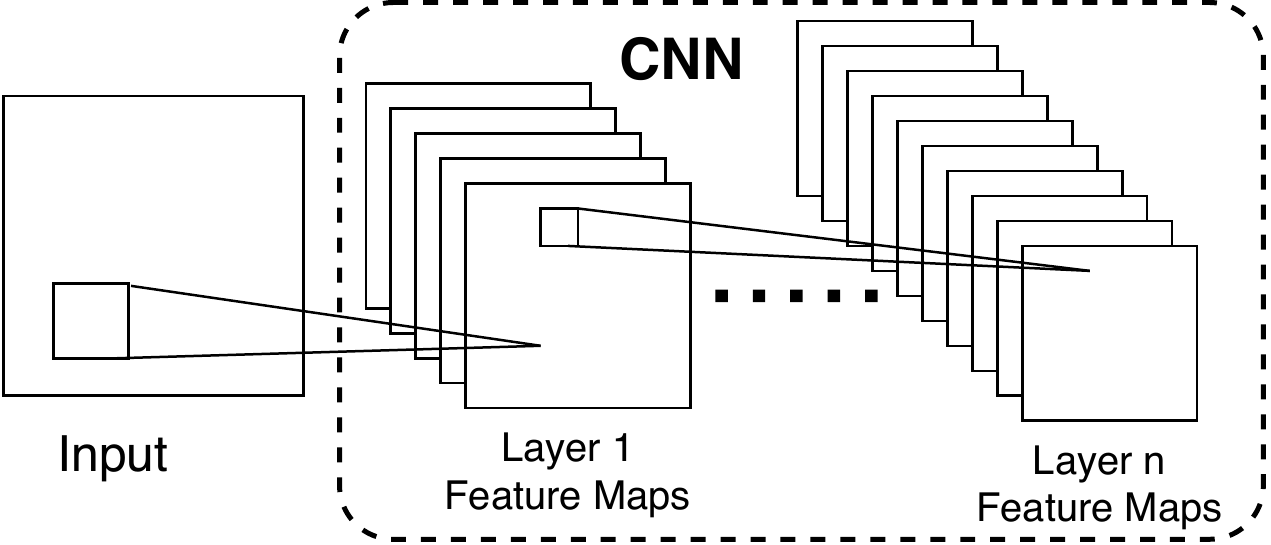}
    \vspace{-7pt}
    \caption{Convolutional feature maps in a classical CNN.}
    \vspace{-22pt}
    \label{fig:cnn}
\end{figure}


With 3D points cloud data, the details for the location of each point are used to calculate the relationship between a car and its surrounding environment. Each frame in 3D points cloud data is processed in the same way, and one common key step in the process is to generate feature maps from points cloud data. Due to the popularity of CNN based solutions to object detection for autonomous driving vehicles \cite{rajaram2016refinenet,li2019stereo,chen2016monocular}, in this paper, we focus on the feature maps generated by the convolutional layers in CNN networks.

\begin{figure*}[!t]
    \centering
    \includegraphics[width=0.9\linewidth]{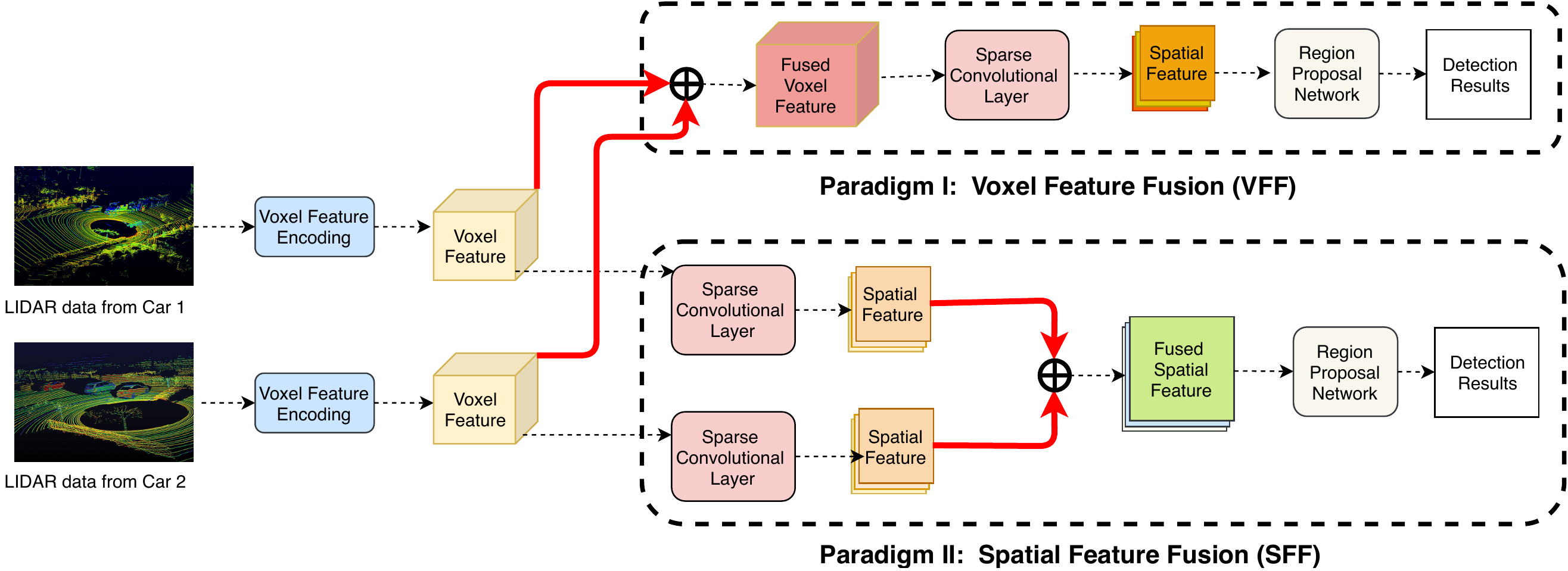}
    \caption{
    Architecture of the feature based cooperative perception (F-Cooper). F-Cooper has multiple vehicles' (using two here for illustration) LiDAR data inputs which are processed by the VFE layers respectively to generate voxel features. To fuse 3D features from two cars, two fusion paradigms are designed: voxel features fusion and spatial features fusion. In Paradigm I, two sets of voxel features are fused first and then spatial feature maps are generated. In Paradigm II, spatial features are first obtained locally on individual vehicles and then fused together to generate the ultimate feature maps. Symbol $\bigoplus$ indicates where the fusion takes place in each paradigm. An RPN is employed for object detection on the ultimate feature maps in both paradigms. We use dashed arrows to denote data flow and bold red lines to present fusion connections. Best viewed in color.}
    \label{fig:architecture}
\end{figure*}

As a CNN network processes raw 3D points cloud data~\cite{ren2017object}, we are able to extract the processed feature maps from the CNN. 
These feature maps provide the essential information for object detection. Fig.~\ref{fig:cnn}  depicts the convolutional layers in a classical CNN. First, we send the original data as input to a convolutional layer which is composed of several filters with each filter generating a feature map. All these generated feature maps are considered as the output of the first layer and will be sent to the second convolutional layer as input data. Recursively, previous layer's outputs are fed as input into the next layer. 



\subsection{\textbf{Features for Fusion}}


Features are a well established and integrated part of any CNN, and due to the nature of CNN, it is opaque by nature. When working with feature maps, we need to ensure that coincident issues are dealt with and explored. 
For example, depending on the specifications of the convolutional layers in a CNN network, the resulting voxel features may be located equal-distant from other voxels, making lossless fusion impossible to achieve without additional run-time cost. 

To confirm the usefulness of features for fusion, we must answer the following three essential questions. (1) Do features possess the necessary means for fusion? (2) Are we able to communicate the data between autonomous vehicles effectively through features? (3) If features satisfy both of the two prior requirements, then how hard is it for us to obtain feature maps from autonomous vehicles?
To analyse these questions and their implications, we provide an in-depth analysis in the sections below.

\subsubsection{\textbf{Fusion Characteristics}}
%

Inspired by the works that have been dedicated to fusing feature maps generated by different layers, such as Feature Pyramid Network (FPN) \cite{lin2017feature} and Cascade R-CNN \cite{cai2018cascade}, we find that it is possible to detect objects in different feature maps. For example, FPN adopts a top-down pyramid structure feature maps for detection. These networks are very adept in compounding the efficiency of feature fusion.



Taking the inspiration from these works, we make the assumption that cars compatible for fusion will use the same detection model.
This is important as we see only the most reliable detection model being used for self driving. With this assumption in place, we now look at the fusion characteristics.



As feature maps are available from the CNN, we are able to ensure that all extracted feature maps are obtained with the same format and data type. Next, as feature maps extracted from 3D points cloud also contain location data, we are able to fuse the feature maps from different autonomous vehicles as long as there exists a single point of overlap in between the two vehicles. 
However, when we faced the issue of equal-distant location alignment, we needed to adjust our fusion algorithm to accommodate such situations. 
To achieve this goal, we let each car send its GPS and IMU data to allow for the transformation calculations towards point clouds fusion, i.e., transforming the view seen by a sender to the view seen by a receiver.
We are clear that GPS and IMU cannot provide enough accurate details to perspective transformation. However, there are existing methods that allow for accurate alignment of two vehicles into the same 3D space. We will discuss this further in the discussion section.



\subsubsection{\textbf{Compression and Transmission}}

Another advantage of feature maps over raw data is the transmission process between vehicles. 
Raw data might come in many different formats, they all achieve a single purpose, and that is to preserve the original state of the data captured. For example, LiDAR data taken from a driving session would store all the points cloud along the path of the driving session. However, this storage format records unnecessary data along with the essential data; \textbf{feature maps avoid this issue}. As the raw data is being processed by the CNN network, all the extraneous data is being filtered out by the network, leaving behind only information that is potentially capable of being used for object detection by the network. These feature maps are stored in sparse matrices, which only store the data deemed useful, with a 0 stored in the matrix for any data filtered out. 

The data size advantage can be further compounded through lossless compression such as the gzip compression method as seen in \cite{gzip}. Adding in the nature of sparse matrix, we are able to combine the two to achieve compressed feature data that is no bigger than 1 MB, making features a great option for deploying On-Edge fusion.

\subsubsection{\textbf{Generic and Inherent Properties}}

All autonomous driving vehicles must base their decisions on the data that their sensors generate. The raw data is generated from the physical sensors on the vehicle before getting forwarded to the onboard computing device. From there, the raw data is fed through a CNN based deep learning network to process the raw data and ultimately make the driving decisions. 

During this process, we are able to pull the extracted features for sharing. By doing so, we are effectively able to obtain the feature maps of the raw data without needing extra computation time or power from the onboard computing device. With the CNN based network being used by almost all known autonomous driving vehicles to date, the feature extraction is generic and does not require further processing before fusion. 


Thanks to the means by which autonomous vehicles process data, we are able to directly take the processed feature maps from the raw LiDAR points cloud data for the purpose of fusion, as this inherently provides location data. As long as the LiDAR sensor has been calibrated to the standards needed for autonomous driving, then we should have a feature map that is capable of retaining the relative locations of all things in relation to the vehicle.
%



\section{F-Cooper: Feature based Cooperative Perception}


Inspired by the advantages of feature map fusion in 2D object detection and the feature maps generated by 3D object detection based on LiDAR data, we propose the Feature based Cooperative Perception (F-Cooper) framework for 3D object detection. Our F-Cooper fuses feature maps generated from two LiDAR data sources oriented in two different aspects. Fusing feature maps (rather than raw data) will not only address privacy issues, but also greatly reduce the network bandwidth requirement.
In F-Cooper, we present two schemes for feature fusion: Voxel Feature Fusion (VFF) and Spatial Feature Fusion (SFF). 
As shown in Fig.~\ref{fig:architecture}, the first scheme directly fuses the feature maps generated by the Voxel Feature Encoding (VFE) layer, while the second scheme fuses the output spatial feature maps generated by the Feature Learning Network (FLN)~\cite{zhou2018voxelnet}.
SFF can be viewed as an enhanced version of VFE, i.e., SFF extracts spatial features locally from voxel features available on individual vehicles before they are transmitted into the network.


\subsection{\textbf{Voxel Features Fusion}}
As with pixels in a bitmap, a voxel represents a value on a regular cube in three-dimensional space.
Within a voxel, there may be zero or several points cloud generated by a LiDAR sensor.
For any voxel containing at least one point, a voxel feature can be generated by the VFE layer of VoxelNet~\cite{zhou2018voxelnet}.

\vspace{-11pt}
\begin{figure}[!h]
    \centering
    \includegraphics[trim={0 0.2cm 0 0.4cm},clip, width=0.9\linewidth]{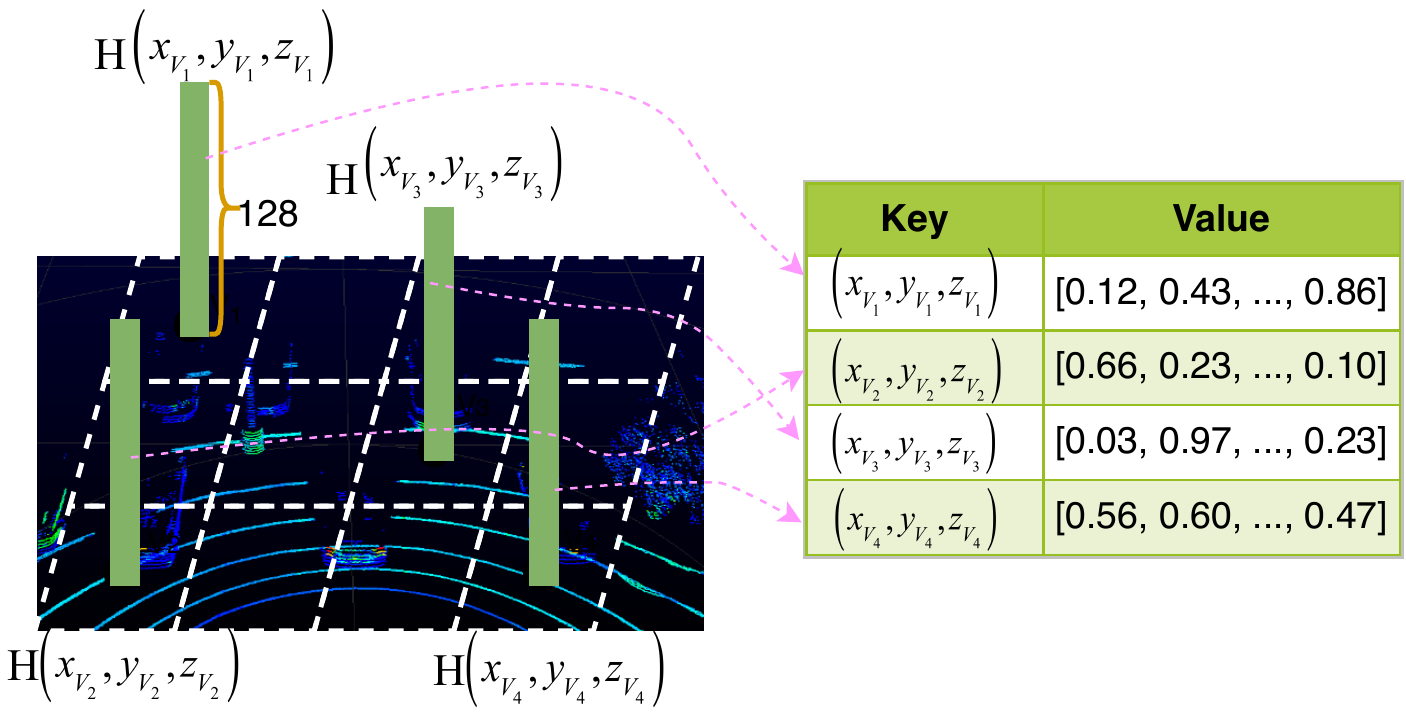}
    \vspace{-7pt}
    \caption{A 128-dimensional feature is generated for each non-empty voxel in LiDAR data. For computational efficiency and data balance, we randomly sample 35 points from the voxels containing more than 35 points. The points within a voxel are then provided to the Voxel Feature Encoding (VFE) layer which produces a 128-dimensional vector. An empty voxel containing no points has no feature.}
    \vspace{-9pt}
    \label{fig:voxel_feature_maps}
\end{figure}

Suppose the original LiDAR detection area is divided into a voxels grid. Of these voxels, we will obtain a vast majority of empty voxels with the remaining ones containing critical information.
All non-empty voxels are transformed by a series of full connection layers and converted into a fixed-size vector, with a length of 128. 
The fixed-sized vector is often referred to as feature map.
An example feature map derived from 3D point cloud data is shown on the right part of Fig.~\ref{fig:3d_feature_maps}.
For example, as shown in Fig.~\ref{fig:voxel_feature_maps}, only four voxels are non-empty amongst the twelve voxels present, and each of the four selected voxels becomes a 128-dimensional vector.

\vspace{-11pt}
\begin{figure}[!h]
    \centering
    \includegraphics[width=0.9\linewidth]{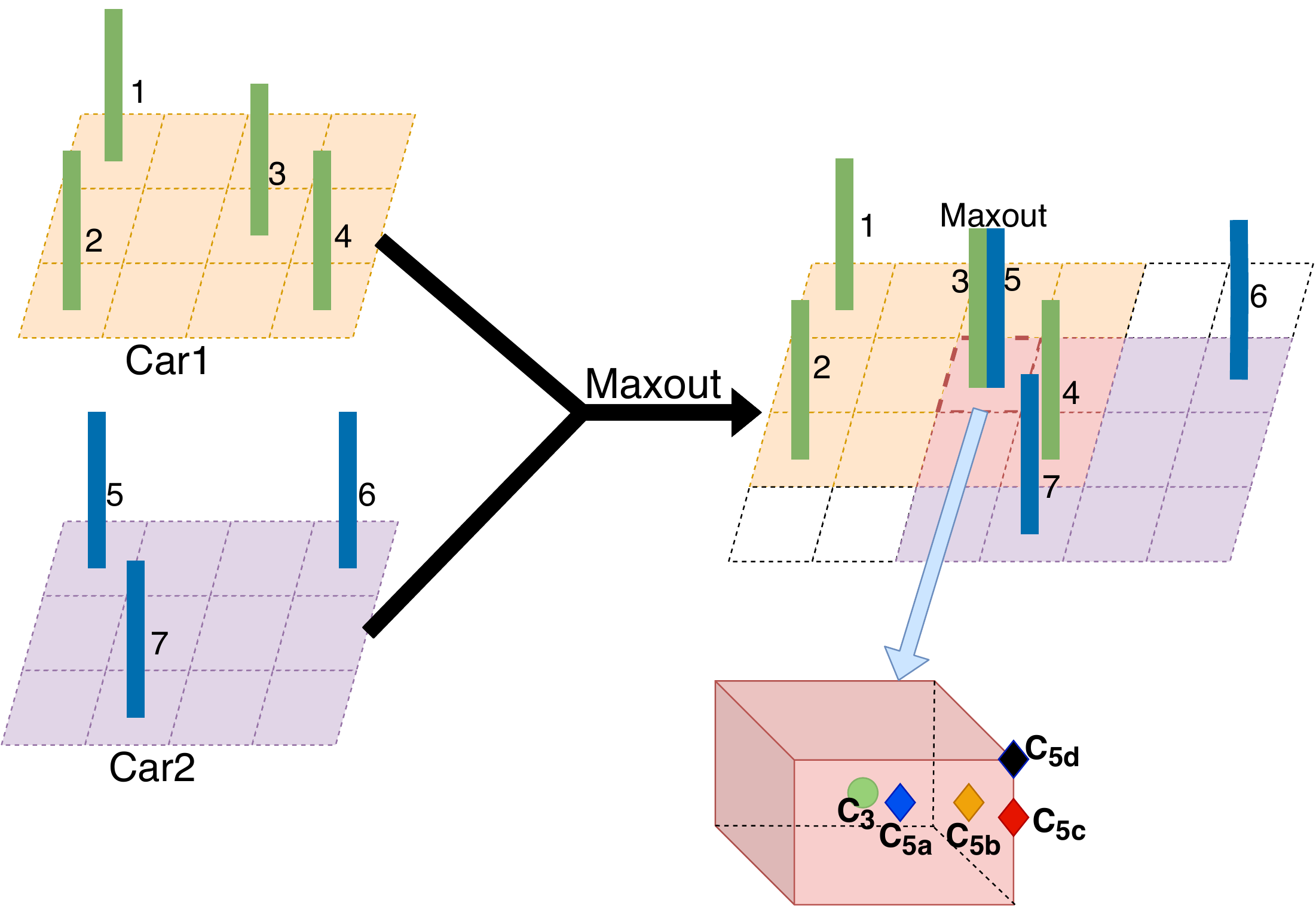}
    \caption{Voxel features fusion. When two voxels share the same location, we use $maxout$ function to fuse them. }
    \vspace{-11pt}
    \label{fig:voxel_fusion}
\end{figure}

For memory/compute efficiency, we save the features of non-empty voxels into a hash table where the voxel coordinates are used as the hash keys. As our focus is primarily on autonomous driving, we only store non-empty voxels into our hash table. Combining the fact that our 3D point clouds are of outside driving scenarios, which yields around a few thousand voxels, searching the hash table for voxel fusion becomes an non-factor in the overall speed of our framework. 
In VFF, we explicitly combine the features of all voxels from two inputs, as depicted in Fig.~\ref{fig:voxel_fusion}.
Specifically, the Voxel $3$ from Car 1 and Voxel $5$ from Car 2 share the same calibrated location. 
While the two cars are located in different locations physically, they share the same calibrated 3D space, with different offsets indicating the relative physical location of each car in said 3D calibrated space.
To this end, we employ the element-wise $maxout$ scheme to fuse Voxel 3 and Voxel 5.

Taking inspiration from convolutional neural networks, using maxout \cite{goodfellow2013maxout} for latent scale selection, we extract the obvious features while suppressing the features that does not contribute to
detection in 3D space, thus achieving lower data size. In our experiments, we use the
$maxout$ to decide which feature is most prominent when comparing the data in between vehicles.
We denote these two voxel features as $V_3$ and $V_5$, respectively, and $V^i$ as the $i$-th element in the voxel. 
The fused features $V$ can be presented as follows.
\begin{equation}
    V^i=max\left ( V_{3}^{i},V_{5}^{i} \right ), \forall i=1,\cdots ,128 
\end{equation}

The key insight behind our $maxout$ strategy is that it emphasizes important features and removes trivial ones. 
Also, as $maxout$  is a simple floating-point operation, it introduces no extra parameters. %
Such a negligible additional computational overhead can be ignored when compared to the overall improvement on object detection precision. 


Naturally, we expect voxels from two cars are able to be perfectly matched. However, this is impractical for real-world applications, even slight bias between voxels would explicitly lead to mismatches. 
Here, we showcase four different mismatched situations in Fig.~\ref{fig:voxel_fusion}.
The green dot $C_3$ indicates the center of the voxel 3 from Car 1 and the diamonds $C_{5a}, C_{5b}, C_{5c}, C_{5d}$ denote the possible centers of the voxel 5 from Car 2.
In case (a), the center of Voxel 5, denoted as $C_{5a}$, falls within Voxel 3. 
In case (b), the center $C_{5b}$ falls on one side of the voxel 3, meaning Voxel 5 connects with two voxels from Car 1. 
In case (c), $C_{5c}$ falls along an edge of Voxel 3, which implies Voxel 5 intersects with four voxels from Car 1.
In case (d), $C_{5d}$ falls on a corner point of Voxel 3 and connects with eight voxels. 
For case (a), we fuse the voxel 3 and voxel 5 directly using maxout. 
For cases (b,c,d), we fuse Voxel 5 with all the connected voxels from Car 1, and give the results to the connected voxels, respectively.


\subsection{\textbf{Spatial Feature Fusion}}
VFF needs to consider the features of all voxels from two  cars, which involves a large amount of data exchanged between vehicles.
To further reduce the network traffic, as well as keeping the benefits of feature based fusion, we design a spatial feature fusion (SFF) scheme.
Compared to VFF, SFF fuses spatial feature maps, which are sparser when compared to voxel features and thus more easily compressed for communication.
Fig.~\ref{fig:architecture} intuitively showcases the relationship between VFF and SFF.
Different from VFF, we pre-process the voxel features on each vehicle to get the spatial features. Next we fuse the two source spatial features together and forward the fused spatial features to a RPN for region proposal and object detection.

\begin{figure}[h]
    \centering
    \includegraphics[width=0.9\linewidth]{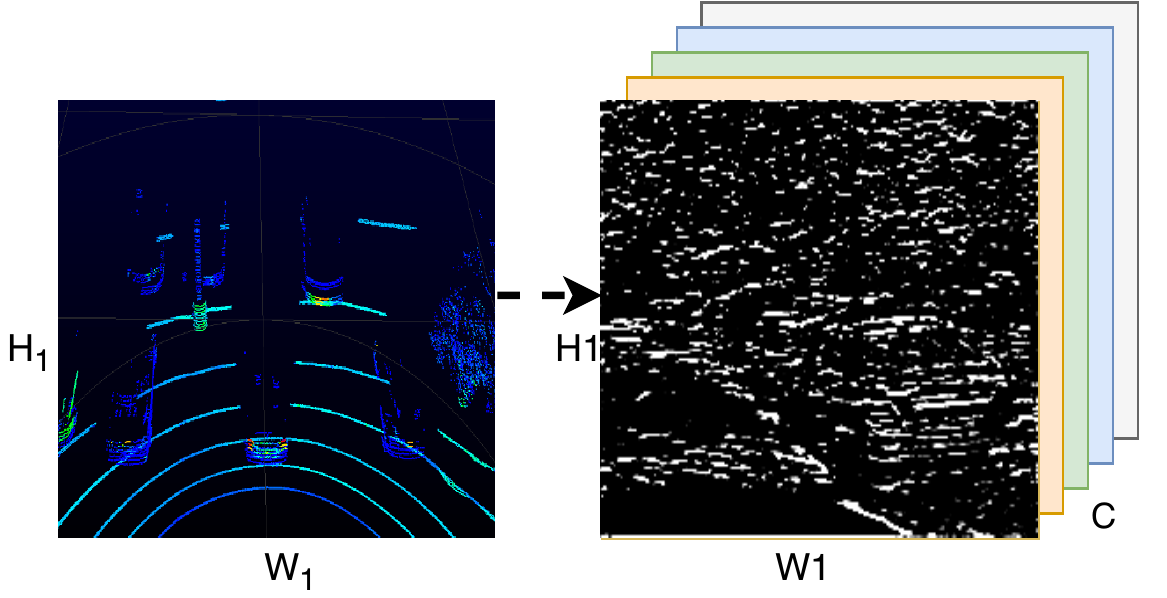}
    \vspace{-7pt}
    \caption{Example of spatial feature maps. $H_1$ and $W_1$ represent the size of the LiDAR bird-eye view for each vehicle's detection range, while $C$ indicates the channels number. It is worth noting that we fuse spatial features in a channel-wise manner, where the channels indicate the corresponding kernel numbers used in CNN.}
    \vspace{-11pt}
    \label{fig:3d_feature_maps}
\end{figure}

As shown in Fig.~\ref{fig:3d_feature_maps}, the spatial feature maps of a LiDAR frame is generated by the Feature Learning Network~\cite{zhou2018voxelnet}. 
The output of the feature learning network is a sparse tensor, which has a shape of $128 \times 10 \times 400 \times 352$. In order to integrate all the voxel features, we adopt three 3D convolutional layers, and sequentially obtain smaller feature maps with more semantic information and a size of $64 \times 2 \times 400 \times 352$. However, the generated features cannot fit into the required shape of the conventional region proposal network. To this end, we must reshape the outputs to the 3D feature maps of size $128\times400\times352$ before we can forward them to RPN.
For SFF, we generate a bigger detection range with size $W\times H$, where $W>W_1, H>H_1$. Next we fuse the overlapped regions while retaining the original features in the non-overlapped regions. Suppose a GPS records the real-world location of Car 1 as ($x_1,y_1$) and Car 2 as ($x_2,y_2$), then we can get the position of the left-top corner. And if $\left(x_2+H_1,y_2-\frac{W_1}{2}\right)$ belongs to Car 2's feature maps with the left-top corner being representative of the feature maps of Car 1, then it is easy for us to determine the overlapped areas. Similar to VFF adopting the $maxout$ strategy, we also employ $maxout$ for SFF to fuse the overlapped spatial features.  
%
%
%

\begin{figure}[!htbp]
    \centering
    \includegraphics[width=0.9\linewidth]{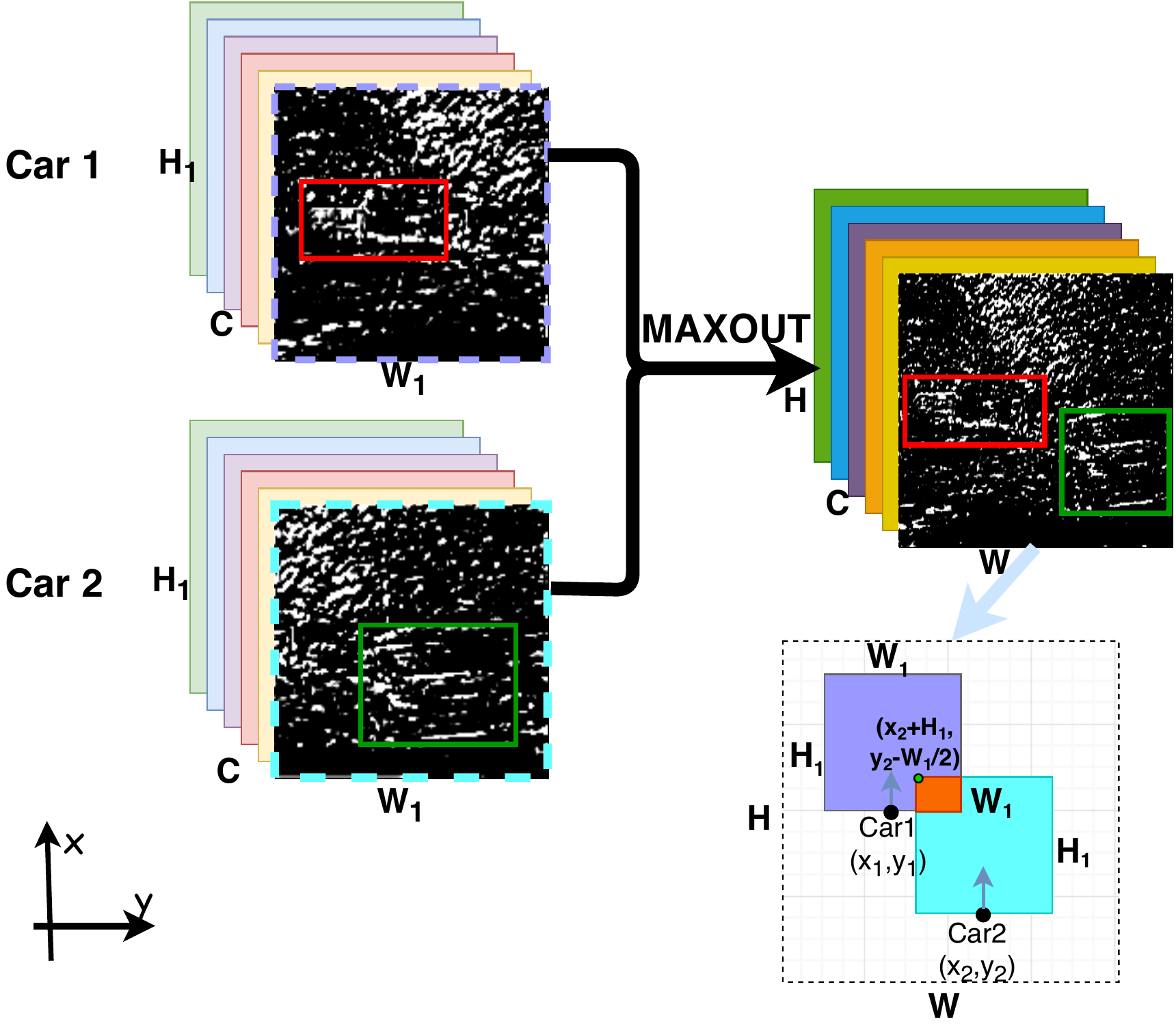}
    \vspace{-7pt}
    \caption{For spatial features fusion, we use $maxout$ to fuse the two spatial features. The left-top is the spatial feature maps generated by Car 1, and the left-bottom is the spatial feature maps generated by Car 2. After fusion, the fused feature maps contain the key features (marked in red and green boxes) of both feature maps.  }
    \vspace{-11pt}
    \label{fig:spatial_fusion}
\end{figure}

As indicated in Fig.~\ref{fig:spatial_fusion}, the top left corner of Car 2's feature maps can be presented as $\left(x_2+H_1,,y_2-\frac{W_1}{2}\right)$ if the car moves towards left.  Suppose the corner point falls in the region of Car 1's feature map, then we can fuse the overlapped features in the same manner as the voxel fusion strategy.  

Finally, we adopt region proposal network to propose potential regions on the fused feature maps. Paradigm II in Fig.~\ref{fig:architecture} holistically showcases the pipeline of our SFF.

Recent work like SENet \cite{hu2018squeeze} indicates that different channels share different weights. That is to say some channels in feature maps contribute more toward classification/detection while other channels being redundant or unneeded. Inspired by this, we opt to select partial channels, out of all 128 channels, to transport. We assume that autonomous vehicles are assembled with the same well-trained detection model as in real-world applications. After extensive experimentation, we demonstrate that transporting part of channels can further reduce the time consumption of transmission while keeping the comparable detection precision in our experimental analysis in Section 4.

\subsection{\textbf{Object Detection Using Fused Features}}
For detecting vehicles, we feed the synthetic feature maps to a Region Propose Network (RPN) for object proposal. Next a loss function is applied for network training. 


\subsubsection{Region Proposed Network}

As indicated in Fig.~\ref{fig:architecture}, once we get the spatial feature maps, regardless of whether we adopt voxel fusion paradigm or spatial fusion paradigm, we send it to the region proposal networks (RPN)  \cite{zhou2018voxelnet}. After passing through the RPN network, we will obtain two generated outputs for a loss function (Section 3.3.2): (1) a probability score $p\in\left [ 0,1 \right ]$ of the proposed region of interests, and (2) the locations of proposed regions $P=\left ( P_x,P_w,P_z,P_l,P_w,P_h,P_\theta  \right )$, where $\left ( P_x,P_y,P_z\right )$  indicates the center of the proposed region and $\left ( P_l,P_w,P_h,P_\theta\right )$ means the length, width, height and rotation angle, respectively.

\subsubsection{Loss Function}
The loss function is comprised of two parts: classification loss $L_{cls}$ and regression loss $L_{reg}$. 


Suppose a 3D ground-truth bounding box can be presented as $G=\left ( G_x,G_y,G_z,G_l,G_w,G_h,G_\theta  \right )$, where $\left ( G_x, G_y, G_z\right )$ represents the central point of the box, and $\left (G_l,G_w,G_h,G_\theta\right)$ denotes the length, width, height and yaw rotation angle, respectively. Our proposed method will generate a vector $P$  to represent the predicted 3D box. In order to minimize the loss between our prediction and the ground truth, we regress our predicted boxes by minimizing the differences $\left ( \Delta x,\Delta y,\Delta z,\Delta l,\Delta w,\Delta \theta   \right )$ \cite{girshick2014rich} as:
\begin{equation}
\begin{split}
\Delta x=\frac{G_x-P_x}{P_d},
\Delta y=\frac{G_y-P_y}{P_d},
\Delta z=\frac{G_z-P_z}{P_h}\\
\Delta l=\log \left ( \frac{G_l}{P_l} \right ),
\Delta w=\log \left ( \frac{G_w}{P_w} \right ),
\Delta h=\log \left ( \frac{G_h}{P_h} \right )\\
\Delta \theta=G_\theta-P_\theta\\
\end{split}
\end{equation}
where $P_d=\left (\left ( P_l \right ) ^2+\left ( P_w \right ) ^2 \right )^{\frac{1}{2}}$ is the dialog of length and width .

Suppose our model proposes $N_{pos}$ positive anchors and $N_{neg}$ negative anchors, we define the loss function as follows:
\begin{equation}
\begin{split}
    L &= \alpha\frac{1}{N_{neg}}\sum_{i=1}^{N_{neg}}L_{cls}\left ( p_{neg}^i ,0\right )\\
    &+ \beta \frac{1}{N_{pos}}\sum_{i=1}^{N_{pos}}L_{cls}\left ( p_{pos}^i ,1\right )\\
    & +\frac{1}{N_{pos}}\sum_{i=1}^{N_{neg}}L_{reg}\left ( P^i ,G^i\right )
\end{split}
\end{equation}
where $p_{neg}^i$ and $p_{pos}^i$ are the probability of positive anchors and negative anchors respectively, and $N_{neg}$ and $N_{pos}$ denote the number of proposed negative and positive anchors respectively. In regression loss, $G^i$ indicates the $i$th ground truth while $P^i$ means the corresponding predicted anchor.  We use $\alpha$ and $\beta$ to balance these three losses. We employ a binary cross entropy loss for classification Loss and Smooth-L1 loss function \cite{girshick2015fast,ren2015faster}.

\section{Performance Evaluation}
\begin{figure*}[!htbp]
\centering
\vspace{-11pt}
\subfloat[Car 1 (Receiver)]{%
\includegraphics[width=55mm,height=100mm]{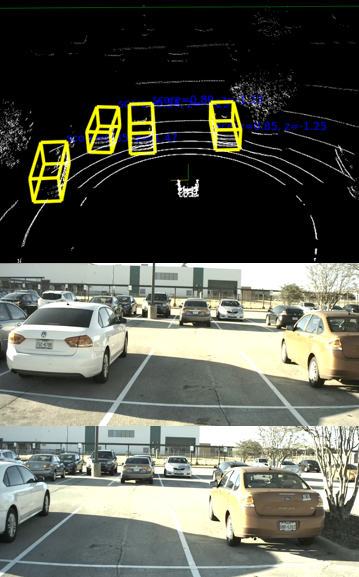}
}
\subfloat[Car 2 (Sender)]{%
\includegraphics[width=55mm,height=100mm]{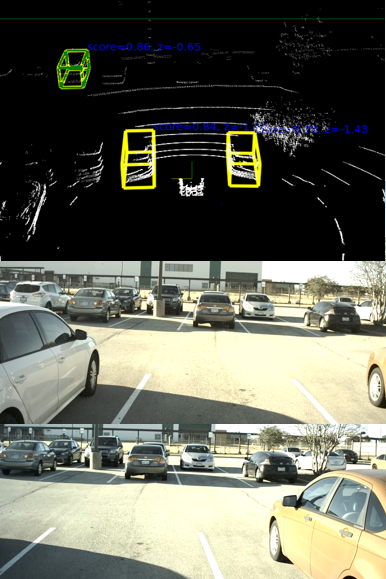}
}
\subfloat[Fusion Detection Result on Car 1]{%
\includegraphics[width=65mm,height=100mm]{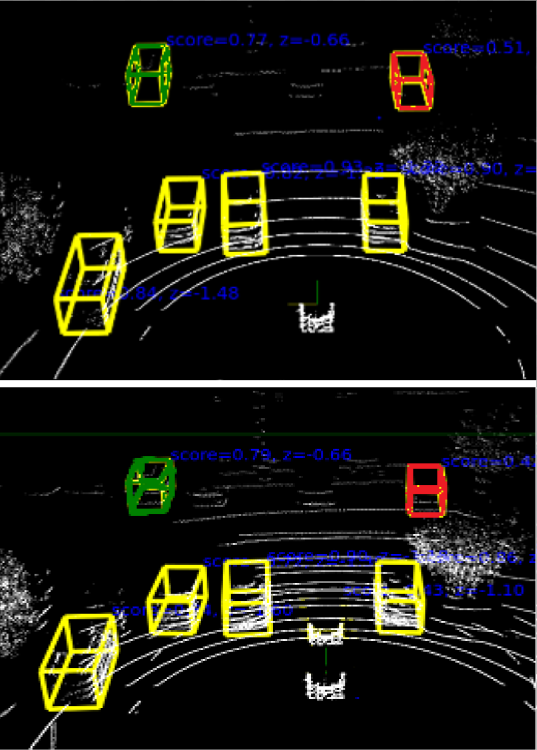}
}
\vfill
\caption{Comparing detection precision on voxel-feature fusion cases when two cars drive forward in parallel . In (a) and (b), the top line is detection results on LiDAR data, while the middle and bottom lines are left and right camera images respectively. In (c), the top line is the result of our voxel fusion and the bottom line is the result of Cooper \cite{qi2019cooper}. }
\vspace{-5pt}
\label{fig:lidar_detect1}
\end{figure*}

\subsection{Datasets}

KITTI \cite{geiger2012we} is a well-known vision benchmark suite project which contains labeled data that allows for autonomous vehicles to train detection models and evaluate detection precision .

%
As we focus on 3D object detection, we use the 3D Velodyne point cloud data provided by the KITTI dataset. The cloud point data provides 100K points per frame and is stored in a binary float matrix. The data includes 3D location of each point and associated reflectance information.
However, as KITTI data is recorded from single vehicles, we must utilize different time segments from the same recording to emulate  data generated from two vehicles. %
As a result, KITTI data is only suitable for certain test scenarios.

To address this issue, we equip two vehicles, named Tom \& Jerry (T\&J), with necessary sensors, such as LiDARs (Velodyne VLP-16), cameras (Allied Vision Mako G-319C), radars (Delphi ESR 2.5), IMU\&GPSes (Xsens MTi-G-710 kit), and edge computing devices (Nvidia Drive PX2) to gather desired data on the campus of our institution. 
Our vehicles have 16-beam Velodyn LiDAR sensors that store  data in binary raw Ethernet packets. As our vehicles can move independently of each other, we are able to test the entire gamut of scenarios in a real-world environment with our two vehicles. 

Both datasets provide data that allows 3D object detection. Moreover, the LiDAR data provided contains ample data for us to extract feature maps from the CNN network.

\subsection{Test Scenarios}

From these two datasets, we are able to fully test or simulate an array of different common scenarios such as those detailed below.

\textit{Road intersections.}
One of the most common places for cars to congregate and thus cause occlusion is a busy road intersection. As the optical based LiDAR and camera sensors are blocked by  vehicles in front of them, the information becomes severely limited. Due to this fact, we included this scenario as one of our test cases.

\textit{Multi-lane roads.}
Another common place  is a multi-lane road. Such roads feature the combination of high speed driving and T-junctions, both of which are prone to accidents. To ensure our F-Cooper framework is capable in such extreme situations, we also included this scenario in our experiments.

\textit{Campus parking lots.}
Last but not least, as our main objective is to enhance perception through fusion, we need to test our framework in a crowded situation with many obstacles. As congested zones are best represented by a crowded parking lot, we choose busy campus parking lots as our main test scenario to evaluate the accuracy of F-Cooper in a real-life environment.

\subsection{Experiment Setup}



To evaluate F-Cooper, over 200 sets of data were collected and tested in our experiments. %
We separate our tests into four categories, based on the methods used to process the LiDAR data, methods (1) through (3) are derived from the same detection model: (1) Non-fusion as baseline, (2) F-Cooper with VFF, (3) F-Cooper with SFF, and (4) raw point clouds fusion method - Cooper \cite{qi2019cooper}.
Feature fusion takes place in random cases of the above four categories with a heavier focus on busy campus parking lots as it is the most difficult scenario due to significant occlusions.
%
%
Within each category, we further divide our experiments by considering the distances between objects and the sensing vehicle.
We treat objects that are within 20 meters away from a vehicle as high-priority objects and those beyond 20 meters as low-priority objects in the parking lot environment. 

As our LiDAR sensor has only 16 beams, the resulting point cloud data is relatively sparse, compared to  higher-end LiDAR sensors. 
To mitigate the negative impacts of sparse data, we limit the detection range to [0,70.4] by [-40,40] by [-3,1] meters along the X, Y, and Z axles.
We do not use data beyond the detection ranges. 
In addition to the vehicle's detection range, we also set the voxel size as ${v_{D}} = 0.4$ meter, ${v_{H}} = 0.2$ meter, ${ v_{W}} = 0.2$ meter, and thus $D_1 = 10$, $H_1 = 400$, and $W_1 = 352$. 
In our experiments, the F-Cooper framework runs on a computer with a GeForce GTX 1080 Ti GPU.

\begin{figure*}[!htbp]
\centering
\vspace{-11pt}
\subfloat[Car 1 (Receiver)]{%
\includegraphics[width=55mm,height=100mm]{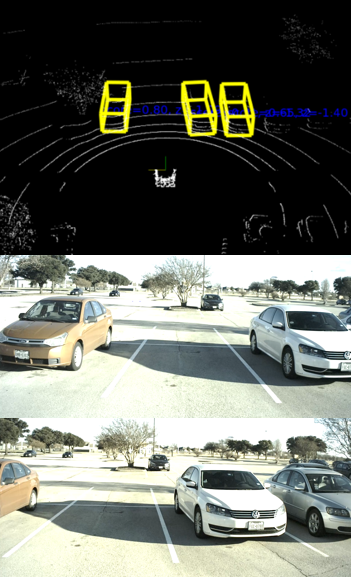}
}
\subfloat[Car 2 (Sender)]{%
\includegraphics[width=55mm,height=100mm]{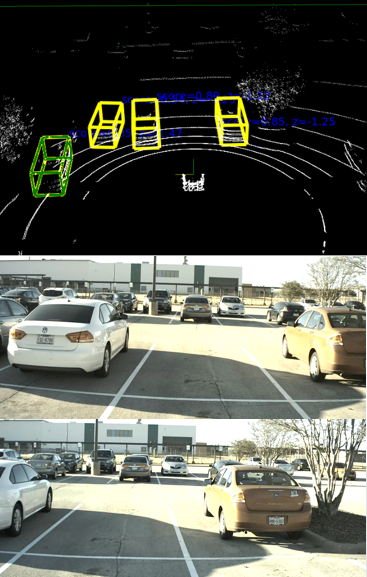}
}
\subfloat[Fusion Detection Result on Car 1]{%
\includegraphics[width=65mm,height=100mm]{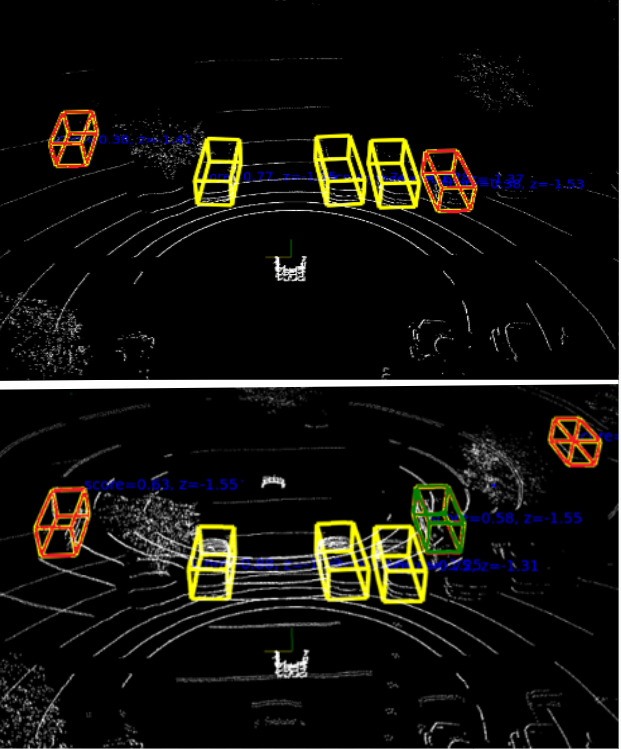}
}
\vfill
\caption{Comparing detection accuracy on spatial-feature fusion cases when two cars approach each other from opposing directions. (a) shows the detection results of car 1, (b) shows the detection results of car 2, while (c) is the results on spatial feature fused data and raw fused data.}
\vspace{-5pt}
\label{fig:lidar_detect2}
\end{figure*}

\subsection{\textbf{Top-Level Evaluation of F-Cooper}}

To evaluate F-Cooper, we analyze each component individually as well as against other frameworks. Starting with VFF, we can see the results of fusion from two cars in Fig.~\ref{fig:lidar_detect1} and Fig.~\ref{fig:lidar_detect2}, with data receiving vehicle (Car 1) and data transmitting vehicle (Car 2). In the figure, we have the LiDAR representation with the detection results on the top and the right-camera in the middle and the left-camera at the bottom. Both the baseline detection and the fusion detection use 0.5 as a confidence threshold, meaning if the confidence level is above this score, we mark the boundary box for that object. As we can see in column (c) of Fig.~\ref{fig:lidar_detect1} and Fig.~\ref{fig:lidar_detect2}, we have the voxel fusion result on the top and the raw data fusion result on the bottom. Through all of our marked bounding boxes, we have distinguished them in three levels of importance to the receiving car: yellow, green and red. The cars marked in yellow represent those that have already been detected by the receiving car originally. The cars marked in green represent those detected by only the sender and not the receiver. The cars marked in red represent those undetected by neither the sender nor the receiver but detected after feature fusion.

Taking a closer look at Fig.~\ref{fig:lidar_detect1}, which details two cars driving forward in parallel, we can see that Car 1 was able to detect four vehicles while Car 2 was able to detect three vehicles. However, in both cases, neither Car 1 nor Car 2 was able to detect cars further away. This was caused by a combination of factors such as occlusion and distance. Through VFF, we are able to detect cars previously occluded to Car 1 or was completely undetected by either cars. 

Similarly, Fig.~\ref{fig:lidar_detect2} depicts two cars approaching each other from opposing directions. In this figure, we can see that Car 1 detects three vehicles while Car 2 detects four. However, when SFF was conducted, we can see that spatial fusion only enhanced perception by two detections for Car 1 where as raw data fusion enhanced detection by three. A closer inspection reveals that the right most new detection from SFF was not detected in the raw data fusion. 
From this comparison, we can see that while VFF is similar in precision to raw data fusion, SFF is able to perform better for near cars when compared to VFF.


\subsection{\textbf{Detection Precision Analysis}}

Having taken an overview of the precision of our two feature fusion methods, we dive into the details of how each method performs against each other as well as against the baseline, and the Cooper approaches~\cite{qi2019cooper}. 

The data that we use for this analysis comes from both datasets to test all of our listed scenarios. 
In all our experiments, we report our results using Intersection over Union (IoU) threshold at 0.7 for vehicles. Then, we calculate the precision by comparing the detected vehicles with the ground truth.

{
\color{red}
\begin{table*}[!thbp]
\centering
\begin{tabular}{|p{25mm}|p{10mm}|p{5mm}|p{5mm}|p{5mm}|p{5mm}|p{5mm}|p{5mm}|p{5mm}|p{5mm}|p{5mm}|p{5mm}|p{5mm}|p{5mm}|p{5mm}|p{5mm}|p{8mm}|}
\hline
Scenario&Dataset&\multicolumn{2}{l|}{Baseline w/o fusion}&\multicolumn{2}{l|}{F-Cooper (VFF)}&\multicolumn{2}{l|}{F-Cooper (SFF)}&\multicolumn{2}{l|}{Cooper \cite{qi2019cooper}}\\
\cline{3-10}
&&Near&Far&Near&Far&Near&Far&Near&Far\\
\hline
Multi-lane roads&KITTI&63.22&22.37&77.46&58.27&50.00&57.14&77.46&71.42\\
\hline
Road Intersections&T\&J &78.37&19.60&80.21&72.37&73.68&53.33&80.21&72.37\\
 \hline
Parking Lot1&T\&J&58.33&33.33&66.67&62.54&66.67&33.33&66.67&70.58\\
\hline
Parking Lot2&T\&J&66.67&18.85&72.25&46.42&72.25&25.00&75.50&50.00\\
\hline
Parking Lot3&T\&J&N/A&45.81&N/A&66.41&N/A&66.41&N/A&66.41\\
\hline
Parking Lot4&T\&J&100&N/A&100&48.83&100&33.33&100&48.83\\
\hline
\end{tabular}
\vspace{10pt}
\caption{Precision comparison between F-Copper and Cooper on Car 1: Average precision (in \%). "N/A" means no vehicle exits in the corresponding scenarios.}
\vspace{-10pt}
\label{table:compare}
\end{table*}
}

In Table \ref{table:compare}, we observe that in our baseline test, baseline without fusion on Car 1  achieve a good ``Near'' detection precision for the road scenarios but fall off sharply in precision in their ``Far'' detection. As mentioned before, the ``Near'' and ``Far'' cut off is $20$ meters from the car as the center. Next, looking at how the baseline performs in parking lot scenarios where the most occlusions take place, we can see that again, the ``Near'' precision is much better than the ``Far''. This is understandable as the lasers reach out further, it returns a much sparser point cloud. 

Moving on to our method testing, we compare the precision against both the baseline and raw fusion~\cite{qi2019cooper}. It should be noted that we only compared against fusion methods instead of non-fusion detection methods as the former yields a meaningful comparison whereas the latter is not relatable in the same context. For our road scenarios, we see that VFF achieves a similar precision to Cooper (a raw data fusion solution). 
\textbf{
This signifies that VFF is as capable as raw data fusion method for near object detection, but without collecting others point clouds.}


Interestingly, as we look at the SFF precision, we can see a drastic difference in between the ``Near'' and ``Far'' precision. While SFF does not outperform VFF, it was still able to perform better than the baselines in most scenarios. However, it must be noted that SFF is more sparse than both voxel features and raw data by a considerable margin. When we factor in the fact that spatial features are derived from the voxel features, it is normal to have the better precision in the regions where the data is naturally denser, i.e., ``Near'' the vehicle where the LiDAR point cloud data is the densest.



\begin{figure}[!h]
\centering
\includegraphics[width=0.33\textheight]{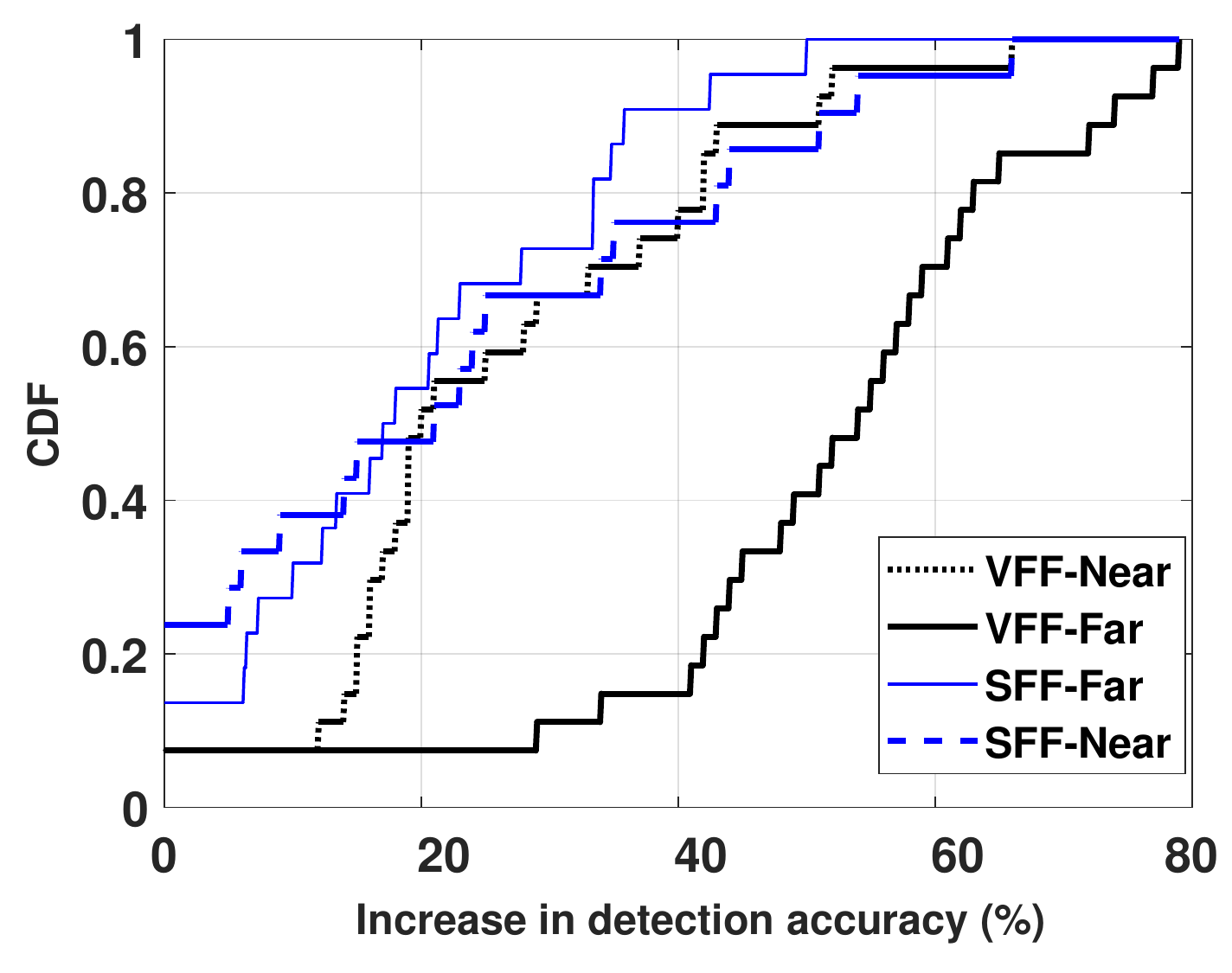}
\vspace{-5pt}
\caption{Cumulative Distribution Function vs. detection precision improvement.}
\vspace{-8pt}
\label{figure:cdf}
\end{figure}

To distinguish the differences in how well VFF and SFFs perform in the ``Near'' and ``Far'' categories respectively, Fig.~\ref{figure:cdf} shows the cumulative distribution functions of increase in detection precision.  
Additionally, in the ``Far'' category, VFF was able to achieve a 40\% detection precision increase for almost 85\% of the time; it is also able to increase detection precision by 60\% for 30\% of the time. Looking at SFF, we do see an increase in detection precision, however, it is not as great of an increase as VFF shows.
When it comes to the ``Near'' category, however, SFF was able to perform as well as VFF if not better in some cases. In Fig.~\ref{figure:cdf}, SFF and VFF are both at 50\% chance to increase detection precision by 20\%. But, as we look deeper, we find that SFF outperforms VFF slightly at 30\% chance to increase precision by 30\%.

We conclude from this test that our detection range is able to be extended with an overall average increase in detection precision due to the extra features being harvested. As our features may target the same object multiple times, the detection confidence scores also see a notable increase. 
The reason why detection results become more precise after fusion is due to the
points from different cars becoming fused, thus making the originally sparse data
representation of a 3D object less sparse and more ``outlined''. This allows for the
detection model to have a higher precision.
Moreover, as single cars have a limited range on their LiDAR beams, multi-car fusion
allows for points in the distance to be registered by the receiving car. Through fusion,
the missing points in the distance are provided by the other cars, and thus allowing for
the recipient car to enhance its detection results.
Our detection precision may increase even further with more vehicles sharing data, solving the issue of missing detection on some of our target cars.

\begin{figure*}[!h]
\centering
\vspace{-11pt}
\subfloat[real-world test]{%
\includegraphics[width=40mm,height=55mm]{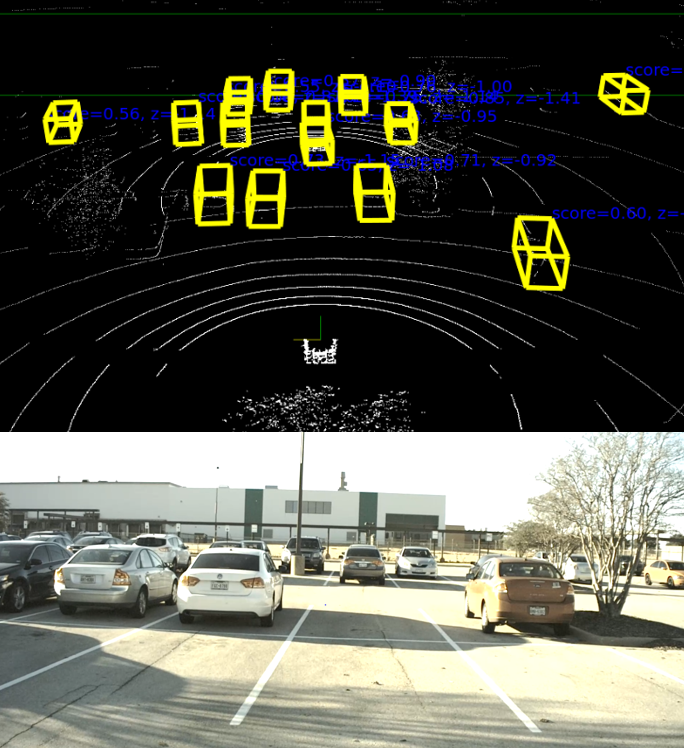}}
\subfloat[VFF detection results]{
\includegraphics[width=65mm,height=55mm]{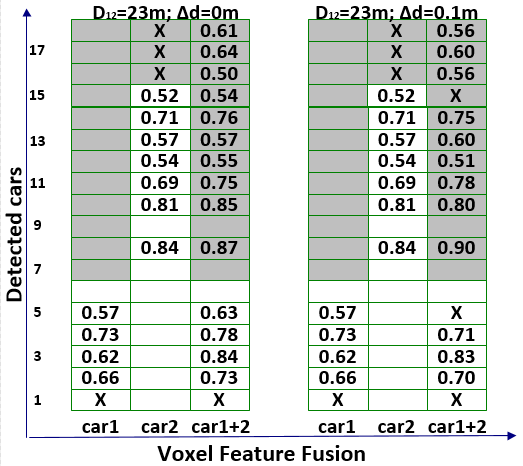}}
\subfloat[SFF detection results]{
\includegraphics[width=65mm,height=55mm]{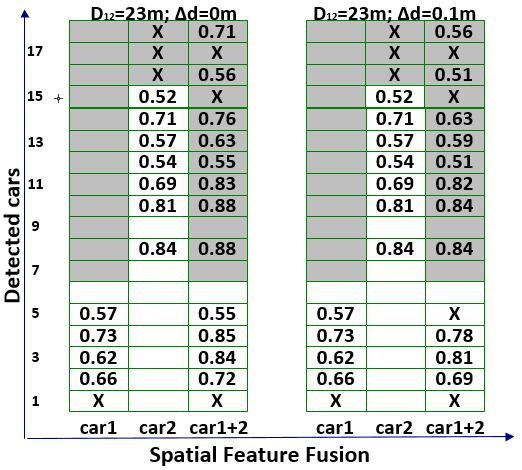}}
\vfill
\caption{GPS reading drifting impact on F-Cooper. (a): intuitive detection result. (b): numeric detection results of VFF. (c): numeric detection results of SFF. The table exhaustively showcases the detection confidence value on each car.}
\vspace{-5pt}
\label{figure:GPS_det}
\end{figure*}

\subsection{\textbf{Sensitivity and Resilience}}

As feature fusion relies heavily on location information for fusion, alignment has a big impact on the final detection precision of the fusion. 
To understand the sensitivity and resilience of F-Cooper, we will not only study missed detections, but also compare the changes of confidence level of each detected target.

In real world situations, all sensors are built within a specific acceptable error tolerance. However, these small discrepancies in between different sensors may cause the same object in 3D space to be labeled at slightly different locations by different cars. As SFF is by nature sensitive to the position of the features, we need to deal with this phenomenon in our fusion. When we integrate our GPS and IMU data, we observe yields of less than 10 cm of positional error \cite{imugps}. Additionally, when we explained the nature of voxel and spatial fusion in Sections 3.1 and 3.2, we noticed the discrepancies in location based data fusion. 
To test the resilience of our fusion methods against sensor drift, we conducted procedural artificial skewing of our GPS readings as seen in Fig~\ref{figure:GPS_det}. 


In Fig.~\ref{figure:GPS_det}, we have part (a) showing the scenarios and part (b) and (c) displaying the effects of GPS drifting on VFF and SFF. First, in both VFF and SFF, we can see that there are two tables, one with a drift of 0 meters and the other with a drift of 0.1 meters to simulate drift. The target cars are then separated into ``Far'' and ``Near'' groups with respect to the location of each vehicle, ``Far'' is shaded dark while ``Near'' is not shaded.

When we focus on the missed detections, the experiment results indicate location drifting does not significantly affect the detection accuracy of SFF.
On the other hand, if we look at the confidence score of each detected target, we find that VFF strategy is not too sensitive to GPS drifting.
%
Taking all of the changes from all of our target scores of VFF, the average of increase and decrease in our confidence scores balance out, indicating that GPS drifting slightly affects VFF. Considering the same scores of SFF, we see that the average of change becomes worse, when compared to VFF, indicating that SFF is more sensitive to GPS drifting than VFF.

During our experimentation, SFF performed worse than VFF in the ``Far'' category. After careful investigation on our experimental setup and methodology, we concluded the following: Compared to raw point cloud fusion and voxel feature fusion, spatial feature fusion is relatively low in feature map resolution. This factor is exponentially amplified during detection for objects in ``Far'' category as well as for detection of small objects.
In retrospect, we realize that for feature extraction on small object, we are even more susceptible to location distance. Furthermore, smaller objects may suffer from missing features after extraction. In point cloud data, when fusing from different angles and perspectives, we are at a higher risk of merging features from different aspects, therein causing a detection conflict. We believe that to overcome this issue, we need to propose a voxel feature extraction method that allows for surgical extraction of features from point clouds.


\subsection{\textbf{F-Cooper On-Edge}}

Even though point clouds can be simplified to coordinate values, we still need to consider the gap between data generated by autonomous vehicles and the limited wireless networking resources, such as the limited bandwidth provided by DSRC.

Due to this limitation, we cannot simply  transmit raw data for the purpose of fusion, as that would congest the network as well as consume valuable on-board computing resources. With F-Cooper, we are able to eliminate this limitation. 

\subsubsection{Transmission}
First, both VFF and SFF are fusion methods that allow for enhanced perception, with VFF achieving close to raw data fusion and SFF achieving better ``Near'' detection results than our baseline. Second, both of our feature fusion methods allow for a final compression size of less than 1 MB, which is well within DSRC limits.

\begin{figure}[!h]
\centering
\includegraphics[width=0.33\textheight]{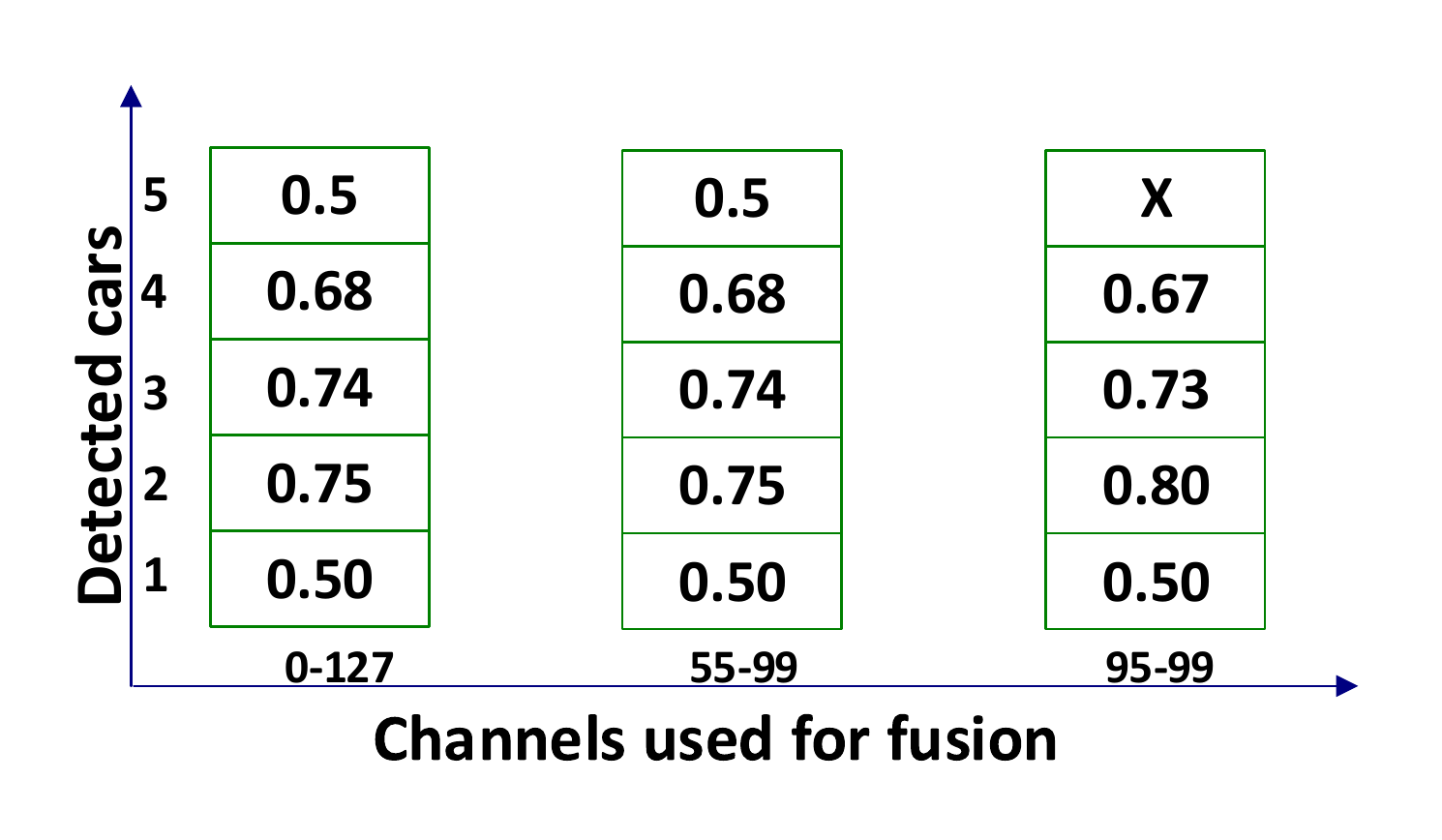}
\vspace{-11pt}
\caption{Detection precision of selective channels on spatial feature fusion. The channels here indicate the corresponding kernel numbers used in CNN.}
\vspace{-11pt}
\label{figure:channel}
\end{figure}

Due to the limitations of DSRC, F-Cooper restricts the frequency of data exchange between vehicles to 1$Hz$ (1 fusion per second). Given the nature of 3D detection and the
situations that we envision, it is not necessary to have a continuous stream of data of
more than 1$Hz$ to achieve enhanced precision. For the majority of cases, only one frame of data is needed to provide crucial supplement to the recipient vehicle. In the case of obstructed views, the feature fusion on a single frame, from different perspectives, will be enough to provide ample warning.

With VFF achieving better results, why do we still need SFF? To answer this question, we analyze the impact of different spatial feature maps on the detection results. As shown in Fig.~\ref{figure:channel}, derived from Fig.~\ref{fig:lidar_detect2}, we have the indexes of channels used in SFF as well as their respective detection precision for each of the 5 vehicles in the scene. 
We have 0-127 channels representing full feature maps usage, while 55-99 channels representing the range of key channels contributing the most to SFF results, 95-99 channels represent a minimal set of required channels to obtain a reasonable detection result.  
\textbf{This finding is crucial for deploying fusion strategies on the edge.} 

\subsubsection{Computation}

\begin{figure}[!h]
\centering
\includegraphics[width=0.33\textheight]{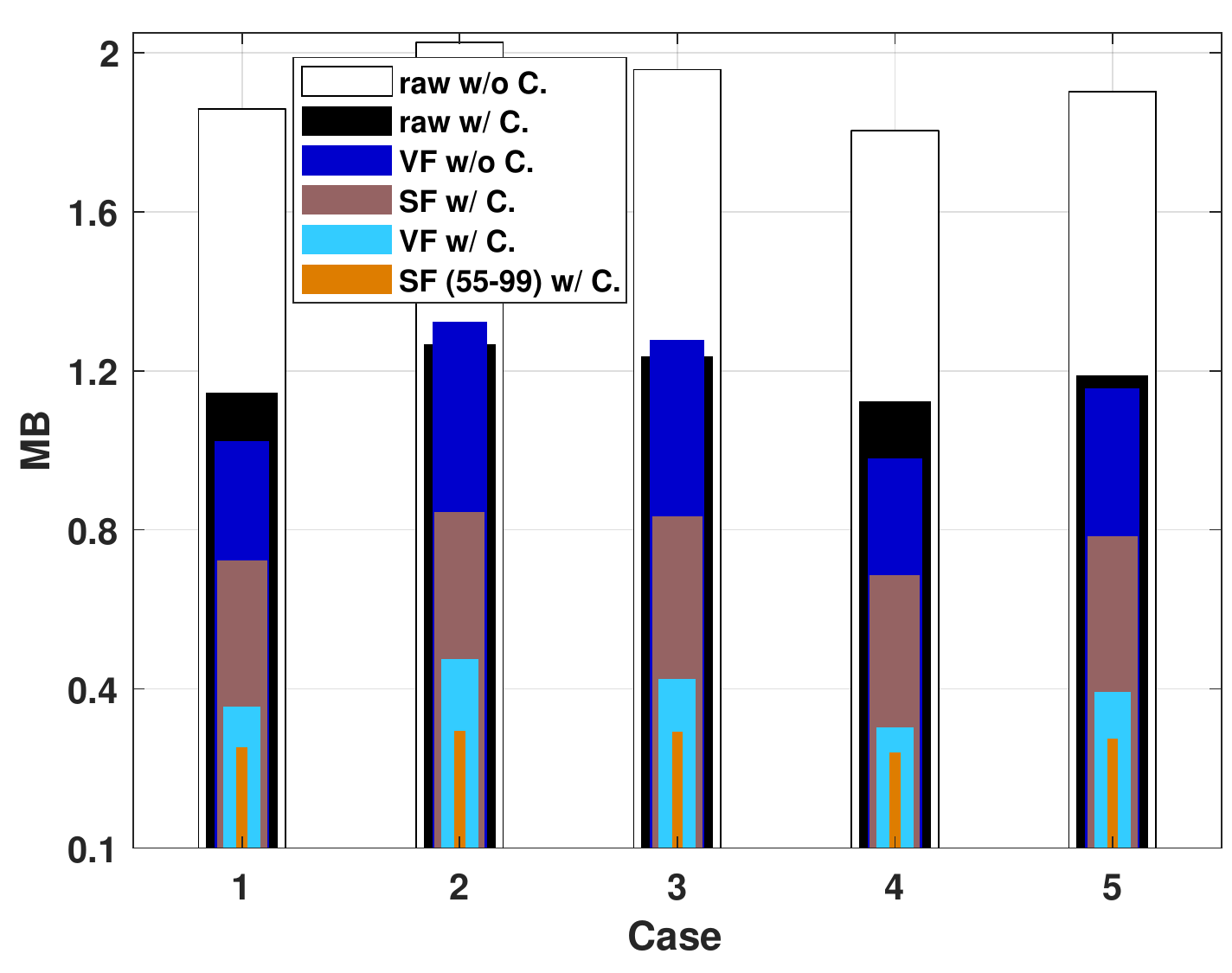}
\vspace{-11pt}
\caption{Comparison (C.) on data volume using different fusion approaches.}
\vspace{-11pt}
\label{figure:size1}
\end{figure}

\begin{figure}[!h]
\vspace{-11pt}
\centering
\includegraphics[width=0.33\textheight]{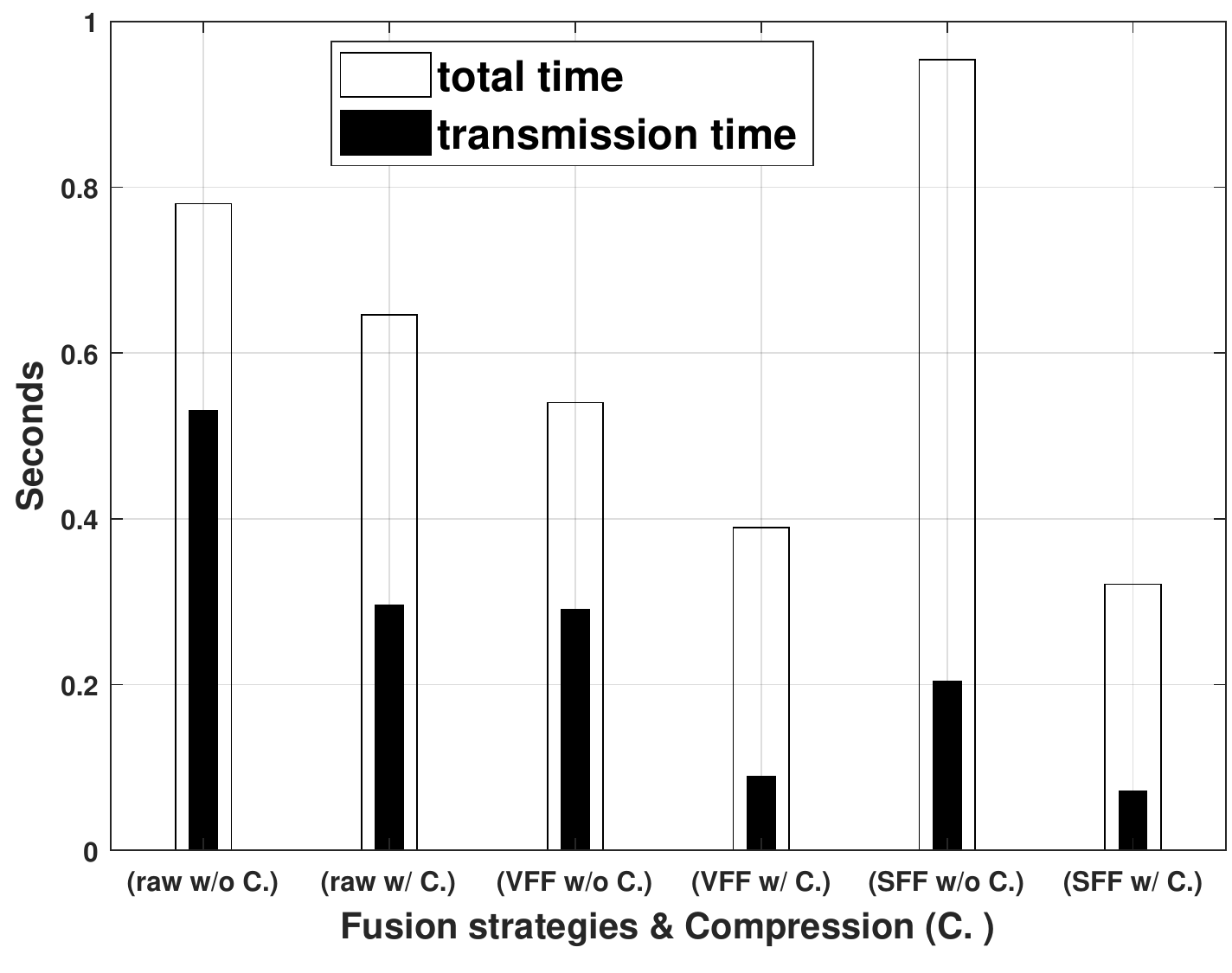}
\vspace{-11pt}
\caption{Comparison on time consuming using different fusion approaches.}
\vspace{-8pt}
\label{figure:size2}
\end{figure}

Due to the small number of channels being used to detect vehicles, we are able to reduce the amount of data that needs to be encoded for compression and transmission. Looking at Fig.~\ref{figure:size1} and Fig.~\ref{figure:size2}, we have the graphs depicting both the data volume and processing time for each of our fusion strategies. 

From Fig.~\ref{figure:size1} we see that the raw point cloud data is about 2 MB when taken directly from our defined LiDAR range as mentioned in the experiment setup section.
Similarly, the original data volume for spatial feature is 72.1 MB and 1 to 1.3 MB for voxel feature. However, both voxel and spatial data is capable of being compressed to less than 1 MB as shown in the figure. When combining the data from Fig.~\ref{figure:channel} and Fig.~\ref{figure:size1}, we can see that with a 55-99 channel SFF compressed, we achieve the highest compression results for all five cases, the average of which is 250 KB. Additionally, if we are to use 95-99 channel SFF, then the end result will achieve an even higher compression. 
At the same time, SFF is capable of achieving a similar precision while being capable of a far better compression. With this, we can now analyze in Fig.~\ref{figure:size2} for how this strategy fares in time consumption. 

Firstly, it should be noted that as vehicles are communicating with each other for data transmission and computation, they are eating up valuable computational resources, so to achieve the best result when it comes to augmenting their perception based on the data from nearby vehicles, edge computing becomes the most important factor.

As shown in Fig.~\ref{figure:size2}, the total time used for both the raw fusion and SFF strategies are both close to the 1 second mark.
Here, the total time we state includes the time for both data processing/transmission and object detection.
This can become quite the issues when compounding this factor with the fact that a single vehicle may need to process the same request from other vehicles at the same time, causing a waste of computational resources. However, when we cut down the total time to just the transmission time needed for the vehicle to transmit and receive the result to and from an edge node, then we have a very feasible method of reliably enhancing perception with no downsides, especially since transmitting features to an edge computing node will not compromise any privacy.

Hence, our fastest strategies only requiring less than a tenth of a second to send and receive results from an edge computing device; the vehicle will only be responsible for sending the data needed for feature fusion without needing  to consume computation resources on decoding, fusing and computing the results from other vehicles.


\subsection{Summary of Experimental Analysis}

We adopted an On-Edge end to end framework, F-Cooper, and achieved a satisfactory collaborative perception towards enhancing detection. Both of our strategies, VFF and SFF, performed better than our single car detection results in almost all of our tests. In addition to better precision, our methods were also lightweight and versatile enough to be deployed in On-Edge systems without adjustments to the current infrastructure of autonomous vehicles.

We also discover that F-Cooper can be leveraged to achieve a reasonable tradeoff in a vehicular edge computing system, considering not only latency and prediction compensation but also data size and network bandwidth.
In our experiments, F-Cooper helps detect more objects that are unclear in the distance. This allows for a less constrained latency range as
the fusion allows for distant objects to be detected before the car in question reaches
that point in space.
In addition, with regards to CNN channel selection and compression, our resulting data sizes make low latency transfers a possibility.
We endeavor to continue researching even more powerful methods in future works.
Lastly, we are only simulating the latency on DSRC channels as that is the most immediate
networking medium. However, there are also 5G and millimeter-wave vehicular communications techniques~\cite{va2016millimeter} coming into play, allowing for much smaller latency.
Latency is a massive issue, and we are not able to solve the real time challenges fully
with our current methodology, but we will continue to strive in our future works.


\section{Related Work}

The exploration of object data fusion has prevailed for years. Usually, data fusion methods can be grouped into 3 categories: low level, feature level and high level data fusion \cite{shi2018leveraging}. 
%

In the era of high level fusion, several works are conducted to fuse the detection results in pursuit of improving detection precision. The work by \cite{rauch2012car2x} exploits a high level sensor data fusion architecture named Car2X-based perception. Their pioneering work delivers one vehicle's consistent results for fusion with the results generated by the host vehicle. High level fusion on multi-sensors has been well investigated to facilitate the development of 3D object detection. \cite{cho2014multi} proposed to detect and track moving objects using fused results from multiple sensors. Recently, Crowd sourcing, which has been learned in an automated manner \cite{qiu2018towards}, has shown competitive perception precision. Sensors from various vehicles are typically crowd-sourced, as cooperators, to provide wider spatial coverage as well as disambiguation.
However, their inability to explore undetected objects and the lack of semantic information communication caused the limited success of cooperative perception system. 
To this end, Qi \textit{et al.} presented Cooper \cite{qi2019cooper}, which fuses original calibrated raw LiDAR data from multiple vehicles to improve 3D detection precision in a low-level data fusion method. 


Though Data fusion has been adopted in many areas, such as object detection and object tracking \cite{luo2018fast}, the idea of fusing data from multiple sources data On-Edge has been explored by only a few authors. An inspiring work is \cite{satyanarayanan2017edge}, where the authors developed a shared real-time situational awareness system by aggregating crowd sourcing and edge computing together.  Another related work is \cite{collaborative}, which employ collaborative learning On-Edge computing. However, the challenges that edge computing needs to face in the specific application of object detection are not mentioned in this paper.


Our fusion strategy is different from previous feature-level data fusion methods. For example, \cite{lin2017feature} fuses features from different convolutional layers in one detection model, \cite{chen2017multi,liang2018deep} fuse features from different sensors within one veihcle. In pursuit of better representation ability, we fuse processed LiDAR features from multiple vehicles. We argue that fusing features from different perspectives is a better solution to improve detection precision. Similar to our work, AVR \cite{qiu2018avr} extends vision of multiple vehicles by communicating short range stereo camera data. The method uses metadata for localization in 3D map, allowing for a much more precise calibration.
Unlike aforementioned works, we present a feature level data fusion method in pursuit of lightweight On-Edge deployment for connected autonomous vehicles.
Our methods are fully suited for On-Edge deployment since the amount of transmitted data is significantly reduced and it effectively takes the advantage of On-Edge computing capacity. Finally, our method is based on intermediate features, which can detect more possible objects than high-level data fusion.

\section{Discussions}

While it is faster to implement high-level data fusion, there is a fundamental flaw associated with this action. As high-level fusion is fusing object detection results from individual cars, we cannot avoid the issue of what if no car senses enough information to detect a critical object. An example would be if car A and car B both detect half of an object, but neither can detect the whole object due to missing half of the point cloud data. Because neither detected the object, the high-level fusion result will exclude the object. 

Another issue involved in data fusion is perspective transformation, in which a receiver needs to estimate its position relative to a sender, so the sender's data can be mapped into the receiver's local coordinate system.
Existing solutions, e.g.,  AVR (augmented vehicular reality)~\cite{qiu2018avr}, have been proposed for precise fusion. 
AVR, with an offline sparse 3D map as the benchmark, can provide an accurate relative localization among vehicles, and thus increase data fusion precision.

Although it is outside the scope of the paper, the fusion of information from different vehicles at the edge opens the door to security vulnerabilities. A prime example can be a malicious vehicle sending phantom vehicle information. This might benefit the malicious vehicle by making space for itself through sending fake information. However, to the general public, this poses a serious driving hazard as they could potentially incur an accident from trying to avoid the phantom vehicle. In addition, we must acknowledge that a vehicle might be unintentionally malicious due to the potential of faulty sensors. This poses the question of how does a vehicle trust the information provided by another vehicle.

Towards these two issues, we assume that all sources are valid and trustworthy for experimentation purposes; however, these issues must be addressed. One possible approach is to have the edge perform the fusion and check the past history of how trustworthy of individual vehicles, and to have the edge perform authentication of newly registered vehicles. 

\section{Conclusions}
In this paper, we proposed F-Cooper, which provides both a new framework for applications On-Edge servicing autonomous vehicles as well as new strategies for 3D fusion detection. Through experiment testing and analysis, we conclude that not only does F-Cooper perform at the same level as Cooper, but it also has the added benefits of being more lightweight and computationally inexpensive. Both voxel features and spatial features have their separate advantages and special uses. Compounded with their great fusion detection enhancing capabilities, both strategies are well suited for autonomous vehicles On-Edge. 

Voxel feature fusion out performs spatial feature fusion, but likewise, spatial feature fusion can be adjusted to be more suited for compression and data transfer. As both methods achieve a high detection perception enhancement over the baseline, both are viable for fusion. 
When we consider the size difference between raw data generated by each autonomous vehicle and only features from the 3D LiDAR data, it becomes clear that the latter is much more suited towards networks with a limited bandwidth. 

When we apply F-Cooper to real-world scenarios, our experimental results on both the data volume and transmission time fall well within acceptable range for On-Edge computation and communication. Thus, from our evaluation, we believe that our proposed F-Cooper framework will add improvement to connected autonomous vehicle system, no matter where or how it is deployed for either in-vehicle or on roadside edge computing.

\section*{Acknowledgment}
The work is supported by National Science Foundation (NSF) grants NSF CNS-1761641 and CNS-1852134. We thank the anonymous reviewers for their many suggestions for improving this paper. In particular we thank our shepherd Ramesh Govindan at the University of Southern California, who have read the previous versions of this paper and provided valuable feedback on
our work.


\bibliographystyle{ACM-Reference-Format}
\bibliography{sample-base}

\end{document}